\documentclass[runningheads]{llncs}
\usepackage{eccv}
\usepackage{eccvabbrv}

\usepackage{tabularx}
\usepackage{booktabs}
\usepackage{caption}
\usepackage{graphicx}
\usepackage{adjustbox}
\usepackage{siunitx}
\usepackage{wrapfig}
\usepackage[table]{xcolor}
\usepackage{xcolor}
\usepackage{amssymb} 
\usepackage{multirow}
\usepackage{colortbl}
\usepackage{tcolorbox}
\usepackage{enumitem}
\usepackage{tikz}
\usepackage{color}
\usepackage[accsupp]{axessibility}
\usepackage[breaklinks,colorlinks]{hyperref}
\usepackage{soul}
\usepackage{amsmath}
\usepackage{subcaption}
\usepackage{pifont}
\usepackage{tocloft}
\usepackage{titletoc}
\usepackage{mathtools}
\usepackage[linesnumbered, commentsnumbered, ruled, vlined]{algorithm2e}
\usepackage{float}
\usepackage{setspace}

\definecolor{molmocolor}{RGB}{240, 82, 156}
\definecolor{tablegray}{RGB}{223, 242, 252}

\sethlcolor{yellow}

\newlength\savewidth\newcommand\shline{\noalign{\global\savewidth\arrayrulewidth
		\global\arrayrulewidth 1pt}\hline\noalign{\global\arrayrulewidth\savewidth}}

\newcommand{\myparagraph}[1]{\vspace{4pt}\noindent\textbf{#1}}

\definecolor{cvprblue}{rgb}{0.21,0.49,0.74}

\newcommand{\circnum}[1]{%
\tikz[baseline=(char.base)]{
\node[shape=circle,draw,fill=white,inner sep=1pt] (char) {\small #1};}
}

\definecolor{lightblue}{RGB}{220, 240, 255}
\newcommand{\crbx}[2]{%
  \scalebox{0.8}{%
    \fcolorbox{#1}{#1}{%
      \raisebox{0.1ex}[1.6ex][0.0ex]{%
        \bfseries\textit{\strut\,#2\,}%
      }%
    }%
  }%
}

\definecolor{IQA}{RGB}{231,236,222}
\definecolor{TIA}{RGB}{254,208,215}
\definecolor{DE}{RGB}{254,205,161}

\definecolor{ODI}{RGB}{234,240,182} 
\definecolor{RC}{RGB}{225,234,207} 
\definecolor{SR}{RGB}{213,227,186} 

\definecolor{FTJ}{RGB}{242,189,193} 
\definecolor{Data}{RGB}{254,202,204} 
\definecolor{HA}{RGB}{248,191,179} 

\newcommand{\IQA}{\crbx{TIA}{\textbf{IQA}}}
\newcommand{\TIA}{\crbx{TIA}{\textbf{T2IA}}}
\newcommand{\DE}{\crbx{TIA}{\textbf{DE}}}

\newcommand{\ODI}{\crbx{ODI}{\textbf{ODI}}}
\newcommand{\RC}{\crbx{ODI}{\textbf{RC}}}
\newcommand{\SR}{\crbx{ODI}{\textbf{3D-SR}}}

\newcommand{\FTJ}{\crbx{DE}{\textbf{FTJ}}}
\newcommand{\Data}{\crbx{DE}{\textbf{Data}}}
\newcommand{\HA}{\crbx{DE}{\textbf{HA}}}

\newcommand*\emptycirc[1][1ex]{\tikz\draw (0,0) circle (#1);} 
\newcommand*\halfcirc[1][1ex]{%
	\begin{tikzpicture}
	\draw[fill] (0,0)-- (90:#1) arc (90:270:#1) -- cycle ;
	\draw (0,0) circle (#1);
	\end{tikzpicture}}
\newcommand*\fullcirc[1][1ex]{\tikz\fill (0,0) circle (#1);}

\definecolor{codegreen}{rgb}{0,0.6,0}
\definecolor{codegray}{rgb}{0.5,0.5,0.5}
\definecolor{codepurple}{rgb}{0.58,0,0.82}
\definecolor{backcolour}{rgb}{0.95,0.95,0.92}

\AtEndDocument{
  \addtocontents{toc}{\protect\par\noindent\rule{\linewidth}{1pt}}%
}

\newcommand{\nocontentsline}[3]{}
\newcommand{\tocless}[2]{\bgroup\let\addcontentsline=\nocontentsline#1{#2}\egroup}

\usepackage{listings}

\lstdefinestyle{mystyle}{
    backgroundcolor=\color{backcolour},   
    commentstyle=\color{codegreen},
    keywordstyle=\color{magenta},
    numberstyle=\tiny\color{codegray},
    stringstyle=\color{codepurple},
    basicstyle=\ttfamily\footnotesize,
    breakatwhitespace=false,         
    breaklines=true,                 
    captionpos=b,                    
    keepspaces=true,                 
    numbers=left,                    
    numbersep=5pt,                  
    showspaces=false,                
    showstringspaces=false,
    showtabs=false,                  
    tabsize=2
}

\definecolor{mblue}{RGB}{0, 77, 128}
\definecolor{darkgreen}{rgb}{0.0, 0.5, 0.0}
\newcommand{\blue}[1]{\textbf{\textcolor{mblue}{#1}}}
\newcommand{\hcolor}[1]{\textbf{\textcolor{darkgreen}{#1}}}

\begin{document}

\title{DynEval: Holistic Evaluations of T2I Generative Models in the Wild} 

\titlerunning{DynEval: Holistic Evaluations of T2I Generative Models in the Wild}

\author{
Shyam Marjit\inst{1}$^{*}$
\and
Dheeraj Baiju\inst{1}$^{*}$
\and
Anuj Shikarkhane\inst{1}$^{*\dagger}$
\and
Akhil Sakthieswaran\inst{1}$^{\dagger}$
\and
Sayak Paul\inst{2}
\and
Anirban Chakraborty\inst{1}
}

\authorrunning{S. Marjit et al.}

\institute{$^1$Indian Institute of Science\quad \quad $^2$Hugging Face \\
\email{shyammarjit@iisc.ac.in, anirban@iisc.ac.in} \\
\vspace{4mm}
Project Page: \url{https://vcl-iisc.github.io/dyneval}
\vspace{-4mm}
}

\maketitle

\begingroup
\renewcommand\thefootnote{}
\footnotetext{* Equal contribution.}
\footnotetext{$\dagger$ Work done during internship at VCL, IISc.}
\endgroup

\begin{abstract}
Recent advances in text-to-image (T2I) generation have led to models capable of producing highly realistic images. Yet, reliably evaluating their outputs remains challenging, especially at scale. Existing automatic evaluators, often relying on a static prompt set, struggle to capture subtle failure modes such as partial prompt misalignment, compositional errors or visually plausible but semantically incorrect generations. In this work, we introduce \textbf{DynEval}, a \underline{\textbf{Dyn}}amic \underline{\textbf{Eval}}uation framework designed to jointly assess \textit{text-to-image alignment} and \textit{image quality} of T2I models. To support scalable training beyond limited human-annotated data, we construct two large datasets. First, we build \textbf{GenDB}, a collection of 500K prompt-image pairs generated from human-written prompts drawn from DiffusionDB using a tiered prompt-model generation strategy. Second, building upon GenDB, we construct \textbf{DynEvalInstruct}, a 250K instruction dataset comprising prompt-image-response triplets distilled from a structured evaluation pipeline that decomposes evaluation into text-image alignment and visual quality reasoning. Using this dataset, we perform full fine-tuning of a compact evaluator through a curriculum learning strategy to effectively distill the superior evaluation capabilities of a larger teacher vision-language model, resulting in \textbf{DynEval-2B} and \textbf{DynEval-4B}. In extensive comparisons against existing evaluators across 11 benchmarks, our evaluator achieves a higher overall correlation with human judgments. Furthermore, it provides fine-grained analysis of the capabilities and failure modes of \textbf{36} T2I models across 42 subcategories and 9 semantic dimensions.
\end{abstract}    
\section{Introduction}
\label{sec:intro}

\begin{figure}[!th]
    \centering
    \includegraphics[width=\linewidth]{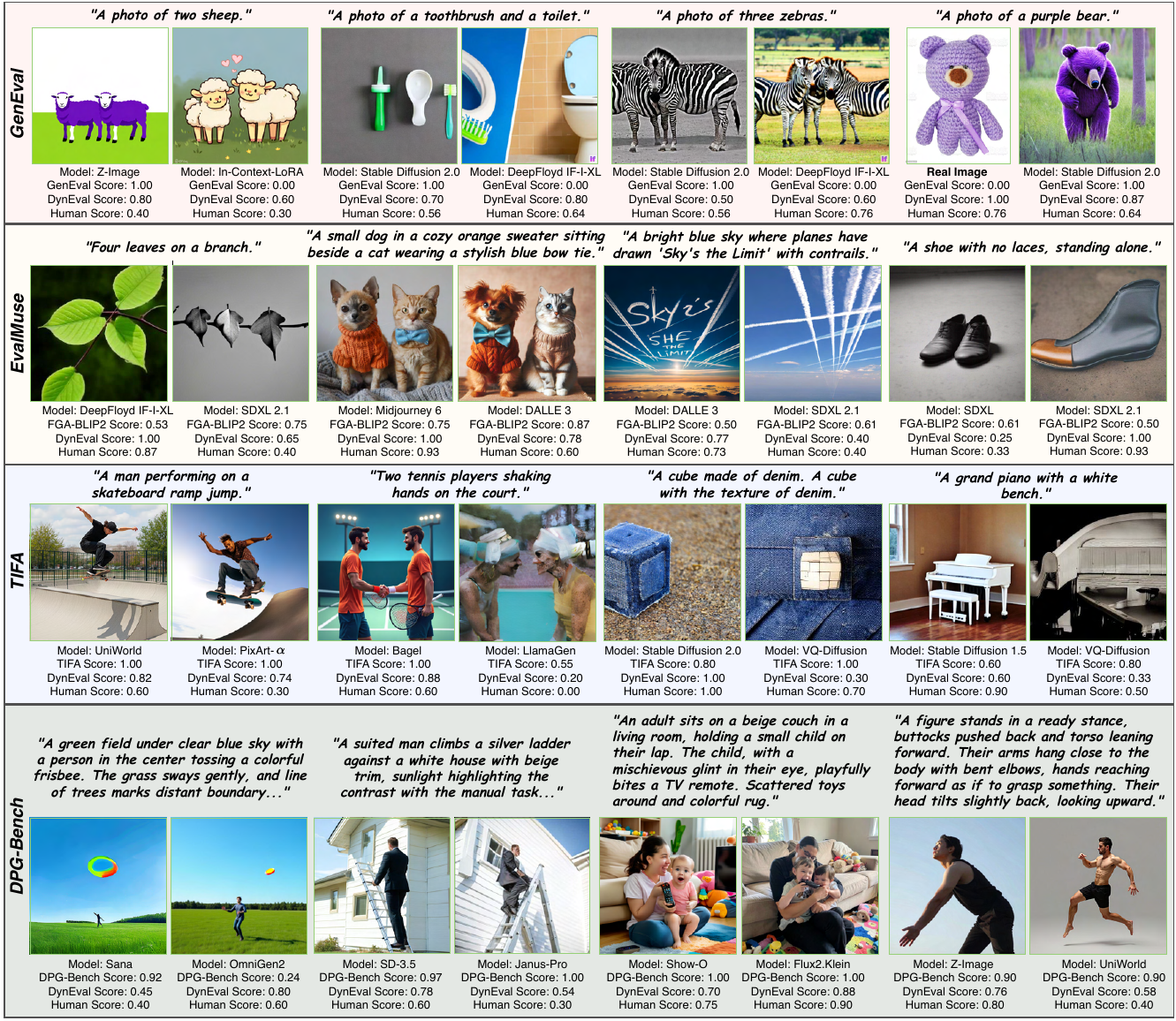}
    \vspace{-7mm}
    \caption{Qualitative comparison of \textbf{DynEval-4B} with four representative text-to-image evaluation methods~\cite{GenEval,tifa,DPGBench,EvalMuse-40k} across multiple T2I models~\cite{rombach2022high,team2025zimage,huang2024context,lin2025uniworld,chen2023pixart,deng2025bagel,sun2024autoregressive,deepfloydif2023,podell2023sdxl,xie2024sana,wu2025omnigen2,esser2024scaling,chen2025janus, xie2024show, flux-2-2025, gu2022vector, betker2023improving, midjourney}. All scores are normalized to the $[0,1]$ range. The first GenEval example additionally includes a real reference image, illustrating the limitations of detector-based evaluation. Compared with existing methods, \textbf{DynEval-4B} produces scores that better agree with human judgments by jointly evaluating text-image alignment and image quality. The examples illustrate three representative failure modes of existing evaluators: \textbf{(i)} Detector-based methods such as GenEval~\cite{GenEval} often fail to accurately evaluate generated content when the outputs exhibit artistic styles or visual distributions that are underrepresented or absent in the detectors' training data; \textbf{(ii)} visually superior images may receive disproportionately low scores; and \textbf{(iii)} visually distorted or semantically implausible images may receive overly high scores. Additional qualitative comparisons are provided in Supp. \cref{fig:supp_teaser_fig}.}
    \label{fig:qualitative_comp}
    \vspace{-6mm}
\end{figure}

Text-to-image (T2I) generation has evolved from multi-stage U-Net~\cite{ronneberger2015u} based diffusion models such as SDXL~\cite{podell2023sdxl} to diffusion transformers~\cite{peebles2023scalable} that scale efficiently to high-fidelity outputs~\cite{esser2024scaling, labs2025flux1kontextflowmatching, flux2024}. In parallel, efficiency-focused diffusion designs have reduced training and deployment costs while enabling high-resolution synthesis on modest hardware~\cite{chen2023pixart, xie2024sana, xie2025sana, chen2025sanasprint, chen2025sana}. Alongside diffusion, auto-regressive generation has emerged as a competing paradigm for visual synthesis~\cite{ sun2024autoregressive, wang2024emu3}. Most recently, unified multi-modal and agentic systems increasingly combine generation with understanding and editing, spanning decoupled-encoder designs, mixture-of-experts reasoning, and hybrid pipelines that emphasize strong text rendering and alignment~\cite{song2025query, cao2025hunyuanimage, chen2025janus, wang2025skywork, deng2025bagel, geng2025x}. With this rapid advancement, evaluating the fine-grained quality and semantic faithfulness of generated content has become a central challenge, as generated images still exhibit T2I misalignment and visual errors (semantic infeasibility, distortion, \emph{etc}.). While human evaluation remains the gold standard for assessing T2I models, it is inherently subjective, expensive, and difficult to scale. Consequently, these limitations have motivated the development of automatic evaluation metrics that aim to mimic human judgment, while remaining scalable.

Early evaluation metrics, including distribution-based measures~\cite{heusel2017gans, binkowski2018demystifying, ramesh2021zero}, CLIP~\cite{radford2021learning}-based metrics~\cite{hessel2021clipscore, park2021benchmark}, and captioning-based variants~\cite{cho2023dall, hinz2020semantic, hong2018inferring, vedantam2015cider, anderson2016spice} offer an incomplete judgment of model performance as they evaluate a limited subset of dimensions and fail to capture many nuanced aspects of image generation. To address this, several structured evaluation methods emerged. Detector-based approaches~\cite{GenEval} verify object-centric attributes but remain sensitive to distortions, texture artifacts, and unrealistic scenes. To further improve semantic assessment, subsequent works have adopted visual question answering (VQA)-based evaluation frameworks~\cite{tifa, tiff_bench, DPGBench, cho2023davidsonian}. However, they depend on proprietary LLMs for decomposing prompts into targeted question-answer (QA) generation and open-source VQA models for scoring, thereby making evaluation costly and difficult to reproduce at scale. Furthermore, approaches that integrate multiple task-specific modules for evaluation~\cite{T2ICompbench} inherit error accumulation due to the integration of different models. To mitigate such error propagation, works like~\cite{T2I-CoReBench, unigenbench++, tiff_bench} rely entirely on closed-source models as evaluators, inherently making evaluation at scale resource-intensive. In contrast, VQAScore~\cite{lin2024vqascore} highlights the utility of open-source vision language models (VLM) as alternatives to proprietary models using the model's probability that an image matches the prompt. However, this formulation can still miss fine-grained distortions and localized compositional failures. To address this limitation, GenEval2~\cite{geneval2} extends VQAScore with Soft-TIFA~\cite{tifa} scoring, but still relies on external LLM generated questions. Interestingly, all of the aforementioned methods lack an end-to-end trained evaluator for generated prompt-image pairs. 

Recent works like EvalMuse~\cite{EvalMuse-40k} and LMM4LMM~\cite{lmm4lmm_evalmi-50k} focus on tuning the scoring VLM using limited ($\approx$ 40-50K) human-annotated ratings on structured questions. While being effective, this human-dependence raises the fundamental question \circnum{1}. Additionally, existing evaluators overlook fine-grained image quality assessment, often producing inconsistent or contradictory results, where low-quality generations receive higher scores while visually superior images are ranked lower (see Fig.~\ref{fig:qualitative_comp}). These observations highlight the need for \circnum{2}.
\vspace{-1mm}
\begin{tcolorbox}[colback=gray!15, colframe=gray!40, boxrule=0.5pt, 
                  arc=2mm, left=3mm, right=3mm, top=1.5mm, bottom=1.5mm]
\circnum{1}\:\textit{Can we train a small-scale VLM-as-a-judge for T2I model evaluation without relying on human ratings, considering the high cost and limited scalability of large-scale human labeling? \\
\circnum{2}\:\textit{Can we build a more reliable and truly dynamic evaluation framework that aligns closely with human judgment, captures real-world perceptual and semantic quality, and adapts dynamically to open-set prompts without relying on external LLMs at inference time?}
}
\end{tcolorbox}
\vspace{-1mm}

\begin{table*}[t]
\centering
\scriptsize
\caption{\textbf{Comparison of capabilities across major T2I-evaluation methods.} Existing methods differ widely in their coverage of key evaluation dimensions such as image quality, object distortions, realism, and dynamic evaluation. Notably, most methods rely on fixed prompt sets and need costly human annotations to train a judge, thereby limiting scalability. In contrast, \textbf{DynEval} introduces a flexible, scalable, and truly dynamic evaluation framework that is independent of a static prompt set. The abbreviations used are: \TIA (\textit{Text-to-Image Alignment}), \IQA (\textit{Image Quality Assessment}), \DE (\textit{Dynamic Evaluation}), \ODI (\textit{Object Distortion Identification}), \RC (\textit{Realism Checking}), \SR (\textit{3D Spatial Relationships}), \FTJ (\textit{Fine-tuned Judge}), \Data (\textit{Tuning Dataset Size}), and \HA (\textit{Human Annotation for Tuning}). Here, \scalebox{0.8}{\halfcirc}, \scalebox{0.8}{\fullcirc}, and \scalebox{0.8}{\emptycirc} denote partial, full, and no support for a capability, respectively.}
\vspace*{-3mm}

\renewcommand{\arraystretch}{1.2}
\setlength\tabcolsep{3pt}
\resizebox{\textwidth}{!}{
\begin{tabular}{r|r|ccc|ccc|ccc}
\shline
\textbf{Method} &
\textbf{Venue} &
\textbf{T2IA} &
\textbf{IQA} &
\textbf{DE} &
\textbf{ODI} &
\textbf{RC} &
\textbf{3D-SR} &
\textbf{FTJ} &
\textbf{Data} &
\textbf{HA} \\
\hline

\textbf{GenEval~\cite{GenEval}}
& \textcolor{gray}{{\tiny NeurIPS'23}}
& \fullcirc
& \emptycirc
& \emptycirc
& \emptycirc
& \emptycirc
& \emptycirc
& \emptycirc
& \emptycirc
& \emptycirc \\

\textbf{TIFA~\cite{tifa}} &
\textcolor{gray}{{\tiny ICCV'23}}
& \fullcirc
& \emptycirc
& \halfcirc
& \emptycirc
& \emptycirc
& \emptycirc
& \emptycirc
& \emptycirc
& \emptycirc \\

\textbf{DPG-Bench~\cite{DPGBench}} & 
\textcolor{gray}{{\tiny arXiv'24}} 
& \fullcirc
& \emptycirc
& \halfcirc
& \emptycirc
& \emptycirc
& \emptycirc
& \emptycirc
& \emptycirc
& \emptycirc \\

\textbf{VQAScore~\cite{lin2024vqascore}} &
\textcolor{gray}{{\tiny ECCV'24}} &
\fullcirc &
\emptycirc &
\fullcirc &
\emptycirc &
\emptycirc &
\emptycirc &
\emptycirc &
\emptycirc &
\emptycirc 
\\

\textbf{TIIF-Bench~\cite{tiff_bench}} &
\textcolor{gray}{{\tiny arXiv'25}} &
\fullcirc &
\halfcirc &
\fullcirc & 
\emptycirc &
\halfcirc &
\halfcirc &
\emptycirc &
\emptycirc &
\emptycirc
\\

\textbf{UniGenBench++~\cite{unigenbench++}} &
\textcolor{gray}{{\tiny arXiv'25}} &
\fullcirc & 
\emptycirc &
\fullcirc &
\emptycirc &
\emptycirc &
\halfcirc & 
\emptycirc &
\emptycirc &
\emptycirc
\\

\textbf{GenEval 2~\cite{geneval2}} &
\textcolor{gray}{{\tiny arXiv'25}} &
\fullcirc &
\emptycirc &
\halfcirc &
\emptycirc &
\emptycirc &
\emptycirc &
\emptycirc &
\emptycirc &
\emptycirc
\\

\textbf{T2I-CompBench++~\cite{T2ICompbench}} &
\textcolor{gray}{{\tiny TPAMI'25}}
& \fullcirc
& \emptycirc
& \halfcirc
& \emptycirc
& \emptycirc
& \fullcirc
& \emptycirc
& \emptycirc
& \emptycirc \\

\textbf{T2I-CoReBench~\cite{T2I-CoReBench}} &
\textcolor{gray}{{\tiny ICLR'26}} &
\fullcirc &
\emptycirc &
\fullcirc &
\emptycirc & 
\emptycirc & 
\emptycirc & 
\emptycirc & 
\emptycirc & 
\emptycirc 
\\

\midrule

\textbf{LMM4LMM~\cite{lmm4lmm_evalmi-50k}} & 
\textcolor{gray}{{\tiny ICCV'25}} &
\fullcirc &
\halfcirc &
\fullcirc &
\emptycirc &
\halfcirc &
\emptycirc & 
\fullcirc &
50K &
\fullcirc 
\\

\textbf{T2I-Eval-Bench~\cite{t2i_eval_bench}} &
\textcolor{gray}{{\tiny ACL'25}} &
\fullcirc &
\halfcirc & 
\fullcirc &
\emptycirc &
\halfcirc &
\emptycirc &
\fullcirc &
14K & 
\emptycirc

\\

\textbf{EvalMuse-40K~\cite{EvalMuse-40k}} &
\textcolor{gray}{{\tiny AAAI'26}} 
& \fullcirc
& \halfcirc
& \halfcirc
& \halfcirc
& \halfcirc
& \emptycirc
& \fullcirc
& 40K
& \fullcirc \\

\textbf{LongT2IBench~\cite{yang2026longt2ibench}} &
\textcolor{gray}{{\tiny AAAI'26}} &
\fullcirc &
\emptycirc &
\fullcirc &
\emptycirc &
\emptycirc &
\emptycirc &
\fullcirc &
14K &
\fullcirc
\\

\rowcolor{cyan!8}
\textbf{DynEval (Ours)} &
\textcolor{gray}{{\tiny ECCV'26}} 
& \fullcirc
& \fullcirc
& \fullcirc
& \fullcirc
& \fullcirc
& \halfcirc
& \fullcirc
& 250K
& \emptycirc \\

\hline
\end{tabular}}
\vspace{-6mm}
\label{tab:benchmark_comparison}
\end{table*}
In this work, we move towards these two goals by introducing \textbf{DynEval}, a \underline{\textbf{Dyn}}amic \underline{\textbf{Eval}}uator for T2I models, obtained through full fine-tuning of a small-scale model using a curriculum learning strategy that effectively distills the evaluation capabilities of a larger VLM in a structured manner. To support this training, we construct two large-scale datasets. First, we curate \textbf{GenDB}, a 500K unique $\langle$prompt, image$\rangle$ pair dataset, where images are generated from human-written prompts sampled from DiffusionDB~\cite{DiffusionDB}. 
Instead of uniformly sampling prompts for image generation, GenDB employs a \textit{tier-matched generation strategy}: prompts are grouped by complexity, T2I models are grouped by capability, and prompts are assigned to models according to their corresponding tiers, ensuring that more capable models are exposed to more challenging prompts. This strategy improves the coverage of informative failure modes across varying levels of prompt complexity and T2I model capability. 
From the curated prompt-image pairs in GenDB, we construct \textbf{DynEvalInstruct}, a dataset of 250K $\langle$prompt, image, response$\rangle$ triplets using the teacher model \texttt{Qwen3-VL-235B}, enabling large-scale supervision through knowledge distillation beyond the scale achievable with manually annotated datasets. Using the DynEvalInstruct dataset, we train compact \textbf{DynEval-2B} and \textbf{DynEval-4B} models as automated evaluators that jointly assess text-to-image alignment (T2IA) and image quality (IQA) for every prompt-image pair. To the best of our knowledge, we are the first to introduce compact evaluators to perform both assessments within a unified framework. For T2IA, DynEval generates prompt-grounded verification questions to determine whether the generated image satisfies the semantic requirements of the prompt. For IQA, it constructs a scene graph from the generated image using the text prompt as contextual guidance, and subsequently leverages this graph to generate image-specific evaluation questions targeting realism, structural consistency, and fine-grained object-level failures.

\begin{figure}[!t]
    \centering
    \includegraphics[width=\linewidth]{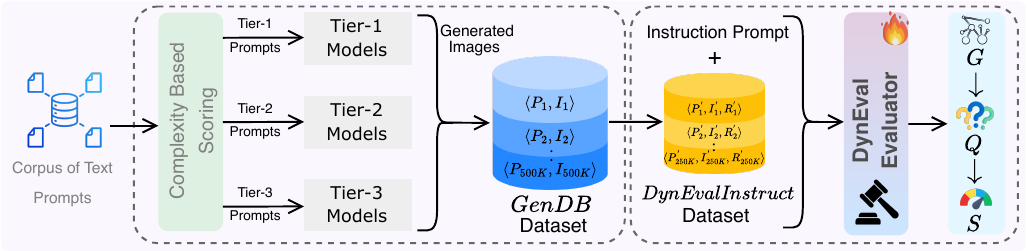}
    \caption{\textbf{Overview of the DynEval Evaluation Framework.}
    \textbf{GenDB} consists of a diverse set of $\langle \text{prompt}, \text{image} \rangle$ pairs (noted as $\langle P_i, I_i \rangle$) obtained by categorizing prompts and models into three tiers (refer to \cref{sec:gendb}), where one image is generated per prompt. DynEval fine-tunes a VLM on the \textbf{DynEvalInstruct} dataset (construction detailed in \cref{sec:DynEvalInstruct}) to generate structured evaluation outputs consisting of T2IA and IQA. T2IA generates prompt-grounded verification questions and answers to assess semantic alignment. IQA is composed of a scene graph ($G$) inferred jointly from the image and prompt and a set of detailed questions ($Q$) for each node and relation in the graph. Across all verification questions generated by the T2IA and IQA processes, we perform VQA-based evaluation and obtain a score $S$ between 1 and 5, assigned to each question.}
    \label{fig:data_methods}
\end{figure}

\noindent The key contributions of this work are summarized as follows:
\begin{itemize}[topsep=0pt, itemsep=0pt, partopsep=0pt, parsep=0pt, leftmargin=*]

\item[$\bullet$] We construct \textbf{GenDB}, a large-scale prompt-image dataset with well-balanced prompt coverage and image generations from \textbf{36} T2I models, and derive \textbf{DynEvalInstruct} from it, an instruction-tuning dataset to train evaluators at scale beyond limited human-annotated data (refer to \cref{fig:data_methods} and \cref{sec:gendb}-\ref{sec:DynEvalInstruct}).

\item[$\bullet$] Unlike static QA methods, we propose \textbf{DynEval}, a dynamic evaluator that jointly evaluates prompt-image alignment as well as builds a scene graph from the generated image to compose structured, image-specific questions for fine-grained image quality assessment (refer to \cref{tab:benchmark_comparison} and \cref{fig:method_instruct}).

\item[$\bullet$] To obtain a robust evaluator, we introduce tier-based prompt categorization with tier-specific image generation to cover a wide gamut of failure modes across varying prompt complexities and model capabilities (refer to \cref{fig:data_methods}).

\item[$\bullet$] We conduct the most comprehensive benchmarking of T2I evaluators to date across \textbf{11} benchmark datasets. Our evaluator achieves higher overall agreement with human judgments (refer to Tabs.~\ref{tab:main_comp}-\ref{tab:added_cross_benchmark_results}), while also providing fine-grained analyses of the capabilities and failure modes of \textbf{36} T2I models across \textbf{42} subcategories spanning \textbf{9} semantic dimensions (refer to \cref{fig:category_score}).

\end{itemize}
\section{Related Work}
\label{main:related_work}

\myparagraph{\textit{Metrics for evaluating T2I models.}} Traditional metrics such as FID~\cite{heusel2017gans}, KID~\cite{binkowski2018demystifying}, and IS~\cite{ramesh2021zero} quantified realism and diversity by comparing the feature distributions of generated and reference images using a pre-trained Inception-V3~\cite{szegedy2016rethinking} model, while LPIPS~\cite{zhang2018perceptual} captured perceptual similarity. However, these image-only measures rely on reference images and assume that class-based features can represent visual realism, making them unsuitable for open-domain, text-conditioned generation~\cite{frolov2021adversarial}. To assess T2I alignment, earlier methods used cosine similarity between DINO~\cite{caron2021emerging} image embeddings and CLIP~\cite{radford2021learning} image-text embeddings (\emph{e.g.}, CLIPScore~\cite{hessel2021clipscore} and CLIP-R~\cite{park2021benchmark}), while captioning-based approaches first convert generated images into textual descriptions~\cite{cho2023dall, hinz2020semantic, hong2018inferring} and compare them to prompts using language metrics such as CIDEr~\cite{vedantam2015cider} and SPICE~\cite{anderson2016spice}. Recent frameworks~\cite{GenEval, T2ICompbench, tifa, DPGBench} further emphasize compositional fidelity through object verification and QA-based evaluation.

\myparagraph{\textit{Object detection based benchmarks.}} T2I evaluation methods often use object detection (\emph{e.g.}, \texttt{Mask2Former}~\cite{mask2former}) to measure compositional fidelity by verifying whether generated images accurately depict prompt elements. Early works such as SOA~\cite{hinz2020semantic} and DALL-Eval~\cite{cho2023dall} check basic properties including object presence, count, and color, while GenEval~\cite{GenEval} further combines object detection with CLIP~\cite{radford2021learning} to evaluate entities, attributes, co-occurrence, and spatial relationships. However, these methods remain limited by the coverage and reliability of the underlying detectors and are generally less suited for abstract prompts, subtle object distortions, global realism, or image-quality failures.

\myparagraph{\textit{VQA-based benchmarks.}} VQA-based evaluation methods use LLMs to decompose prompts into QA pairs and evaluate generated images by performing VQA on these questions, providing a more fine-grained and interpretable measure of prompt faithfulness than CLIP-based metrics. TIFA~\cite{tifa} employs \texttt{GPT-3}~\cite{brown2020language} to generate verifiable QA pairs covering attributes such as color, count, and spatial relations, and then evaluates the generated images using \texttt{mPLUG}~\cite{li2022mplug} or \texttt{BLIP-2}~\cite{li2023blip}. DPG-Bench~\cite{DPGBench} extends this paradigm by leveraging DSG~\cite{cho2023davidsonian} to decompose complex prompts into atomic predicates, followed by VQA with \texttt{mPLUG}~\cite{li2022mplug} for fine-grained assessment of compositional and multi-object semantics. More recently, TIIF-Bench~\cite{tiff_bench} evaluates whether T2I models can preserve the same semantic intent across short and long prompt formulations. GenEval2~\cite{geneval2} addresses benchmark drift in GenEval by replacing binary object-level scoring with Soft-TIFA, a VQAScore-style~\cite{lin2024vqascore} primitive-level evaluator designed to better correlate with human judgments. T2I-Eval-Bench~\cite{t2i_eval_bench} takes a complementary direction by decomposing T2I evaluation into simpler sub-tasks and distilling the resulting evaluation capability into an open-source MLLM. Similarly, T2I-CoReBench~\cite{T2I-CoReBench} separates explicit composition from implicit reasoning: it evaluates generated scene-graph elements (using \texttt{Gemini 2.5 Flash}) such as instances, attributes, and relations, but also probes deductive, inductive, and abductive inference. UniGenBench++~\cite{unigenbench++} broadens semantic evaluation (using \texttt{Gemini 2.5 Flash}) through a hierarchical taxonomy of themes and evaluation criteria, and further tests robustness across English/Chinese and short/long prompt variants. These works demonstrate that question-based decomposition enables fine-grained diagnosis; however, most prior VQA-style protocols primarily focus on prompt satisfaction via discrete semantic checks, often overlooking image quality (\emph{i.e.}, IQA), realism, and structural distortions.

\myparagraph{\textit{Hybrid benchmarks.}} These approaches combine VQA, object detection, and traditional metrics to provide a more holistic assessment of T2I generations. T2I-CompBench++~\cite{T2ICompbench} tailors evaluation metrics to different aspects of compositionality: disentangled BLIP-VQA~\cite{li2022blip} for attribute binding, UniDet~\cite{zhou2022simple}-based evaluation for 2D/3D spatial relationships, and a fine-tuning scheme to strengthen prompt alignment. Similarly, OmniGenBench~\cite{wang2025omnigenbench} adopts a dual-mode evaluation protocol: perception-centric evaluation employs off-the-shelf visual parsing tools, while cognition-centric evaluation leverages \texttt{Gemini-2.0-Flash} as an LLM judge to assess the alignment between generated images and user instructions through task-specific evaluation prompts. Complementing these approaches, HEIM~\cite{lee2023holistic} combines human evaluation and traditional metrics across 12 dimensions, including alignment, aesthetics, bias, and robustness.

\myparagraph{\textit{Human preference guided benchmarks.}} Beyond objective alignment and fidelity metrics, prior works incorporate human preference to capture higher-level qualities such as aesthetics, creativity, and perceptual appeal. ImageReward~\cite{imagereward} trains a reward model from human ratings and pairwise comparisons to rank images according to human preference for alignment and fidelity. VisionReward~\cite{visionreward} learns an interpretable human-preference score by decomposing visual quality into hierarchical dimensions and fine-grained binary checks, followed by learning linear weights from pairwise human preferences. EvalMi-50K~\cite{lmm4lmm_evalmi-50k} and EvalMuse-40K~\cite{EvalMuse-40k} construct human-annotated evaluation datasets spanning perception and text-image correspondence to train the underlying judge model. In contrast, our framework trains the evaluator entirely without human annotations.

\section{\textit{GenDB} Dataset}
\label{sec:gendb}

\begin{figure*}[t]
    \centering
    \includegraphics[width=\linewidth]{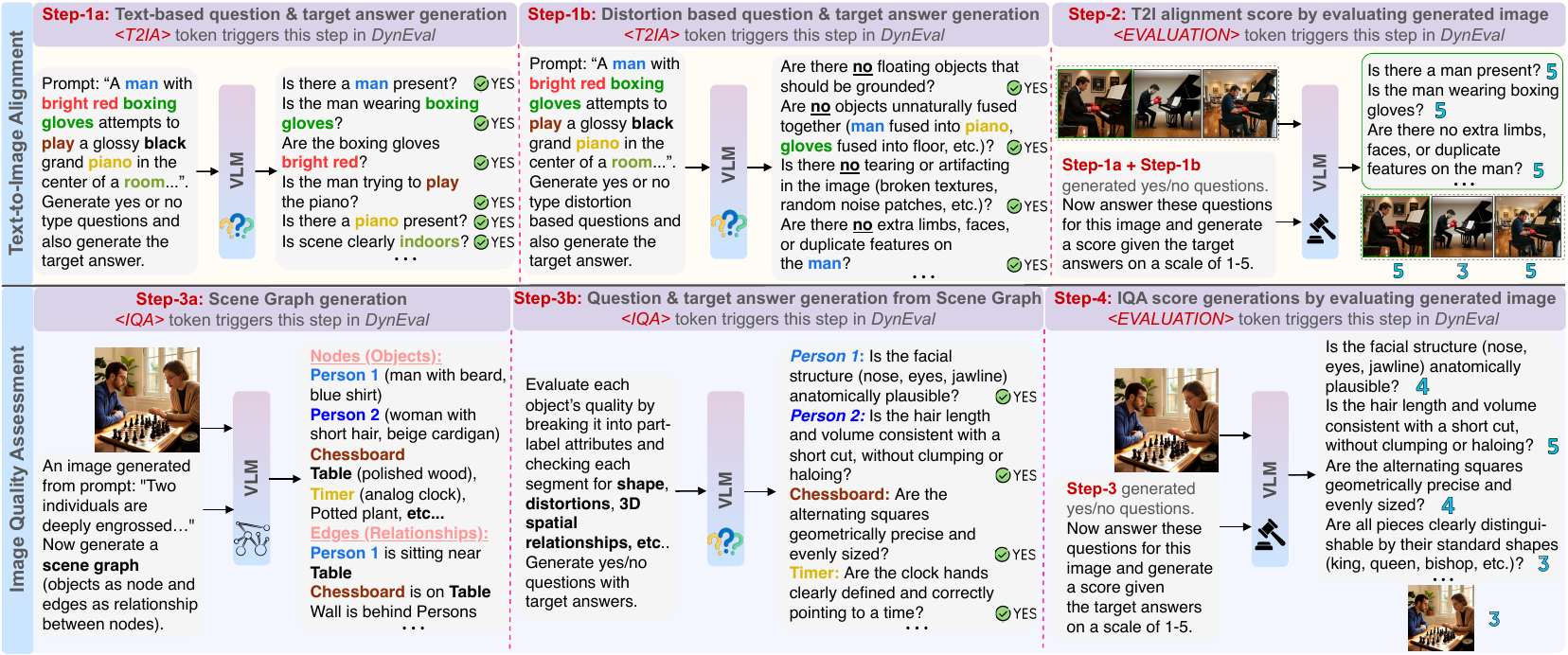}
    \caption{
    \textbf{Overview of the proposed \textbf{DynEvalInstruct} generation pipeline and \textbf{DynEval} training framework.} The teacher VLM (\texttt{Qwen3-VL-235B}~\cite{Bai2025qwen}) decomposes evaluation into two complementary dimensions: \textbf{T2IA}, which generates prompt-grounded semantic and distortion-checking questions and target answers, and \textbf{IQA}, which constructs scene graphs and object-level quality assessment questions and answers. The teacher model then performs VQA-based scoring on the generated question sets to obtain T2IA and IQA assessments. The resulting dataset consists of $\langle$prompt, image, response$\rangle$ triplets, where the \textbf{response} comprises the complete structured T2IA and IQA evaluation outputs generated in \underline{Steps~1--4}, including generated questions, target answers, scene graphs, and question-level evaluation scores. These annotations collectively provide fine-grained supervision for training \textbf{DynEval} with three task-specific tokens: \texttt{<T2IA>}, \texttt{<IQA>}, and \texttt{<EVALUATION>}. DynEval is trained via curriculum learning, first learning to generate structured questions and then to perform question-based evaluation, enabling dynamic and holistic assessment of arbitrary text-image pairs.}
    \label{fig:method_instruct}
\end{figure*}

\myparagraph{\textit{Overview.}} Our goal is to develop a dynamic and robust evaluator that diagnoses T2I model failure modes under \emph{real-world user prompts}. To achieve this, we begin by constructing \textbf{GenDB}, a large-scale dataset designed to capture diverse prompt complexities and model capabilities. We then curate prompts using a complexity-based scoring rule (\emph{prompt tiering}) and stratify 36 T2I models by capability using the \textbf{DynEval-1K} evaluation set (\emph{model tiering}). Finally, we generate 500K $\langle$prompt, image$\rangle$ pairs through tier-matched prompt-model assignment. GenDB serves as the foundation for constructing \textbf{DynEvalInstruct} (Sec.~\ref{sec:DynEvalInstruct}), which is used to train the \textbf{DynEval} models (Sec.~\ref{sec:dyneval}).

\myparagraph{\textit{Complexity-based Scoring.}} We sample prompts from \textbf{DiffusionDB}~\cite{DiffusionDB}, which contains 1.8M unique prompts specified by real users, to better reflect real-world user prompting while avoiding the handcrafted templates and LLM-generated prompts commonly used in existing benchmarks. From this large pool, we curate challenging prompts using a filtering and ranking pipeline. Concretely, each prompt is assigned a \textbf{heuristic complexity score} based on \textbf{9} factors including \textit{prompt length}, \textit{object and attribute counts}, and seven other \textit{semantic attributes}. We discard short prompts and rank the remainder by complexity score. More details on the considered factors and how these factors translate to the complexity score are provided in the Supp. \cref{supp:cbs}.

\myparagraph{\textit{Prompt Tiering.}} We construct a pool of 500K prompts from DiffusionDB~\cite{DiffusionDB} by selecting diverse and challenging prompts using the aforementioned \textit{complexity}-\textit{based scoring} rule. After ranking by complexity, we split the selected 500K prompts into three difficulty tiers using two adaptive score thresholds ($\tau_1$ and $\tau_2$), yielding Tier-1 (hard, score $>\tau_1$), Tier-2 (medium, $\tau_2<$ score $\leq \tau_1$), and Tier-3 (easy, score $\leq \tau_2$) prompts. Details of the thresholding strategy and tier-wise illustrative prompt examples are provided in the Supp. \cref{supp:pt}.

\myparagraph{\textit{Model Tiering.}} Now, to group the 36 T2I models into three capability tiers, we carefully sample 1,000 prompts from the evaluation prompt pools of \cite{EvalMuse-40k, GenEval, DPGBench, tifa}, ensuring uniform coverage across 42 subcategories spanning 9 semantic dimensions (\emph{e.g.}, `Object \& Entity' covers subcategories like \textit{Human} and \textit{Vehicle}; see \cref{sec:failure}). Thereafter, each of these 1,000 prompts is processed by all 36 T2I models, resulting in 1,000 images per model and a total of 36K $\langle$prompt, image$\rangle$ pairs. We refer to this dataset as the \textbf{DynEval-1K} evaluation set, which is also later used for model ablations and failure analysis. Now, each pair from this set of 36K is scored using \texttt{Qwen3-VL-235B}~\cite{Bai2025qwen}, the same large VLM used as the teacher model during distillation. Following our holistic evaluation framework (Fig.~\ref{fig:method_instruct}), the score aggregates two aspects: T2IA and IQA (same strategy as described in detail in \cref{sec:DynEvalInstruct}). We then average scores across the 1,000 prompts to obtain a final score for each model. Finally, we apply two thresholds ($\mu_1 > \mu_2$) to these aggregated scores to partition models into three tiers, where Tier-1 denotes the strongest group and Tier-3 comprises the weakest among the considered models. The exhaustive list of tier assignments is provided in the Supp. \cref{supp:mt}.

\myparagraph{\textit{Tier-matched pairing to construct GenDB.}} Finally, once the T2I models and prompts are grouped into three tiers, we construct \textbf{GenDB}, a dataset of 500K $\langle$prompt, image$\rangle$ pairs. Specifically, we use the previously sampled 500K prompts (from DiffusionDB~\cite{DiffusionDB}) and generate images using 36 T2I models while enforcing tier consistency, \emph{i.e.}, Tier-$i$ models are paired exclusively with Tier-$i$ prompts. To maximize prompt diversity under a fixed computational budget, we do not generate images from all 36 models for each prompt. Instead, within each tier, prompts are randomly partitioned into disjoint subsets and assigned to individual models, ensuring that each prompt is processed by exactly one model. This tiered structure is motivated by the observation that Tier-3 models often perform poorly even on simple prompts, making evaluation on more complex prompts less effective. Conversely, Tier-1 models tend to perform well on simple prompts, yielding fewer failure cases for training a robust evaluator. Therefore, it is necessary to pair models with prompts commensurate with their category.

\section{\textit{DynEvalInstruct} Dataset}
\label{sec:DynEvalInstruct}

\myparagraph{\textit{Overview}.} While constructing GenDB, we employ complexity-based prompt sorting and prompt-model tier matching to increase the likelihood of capturing failure cases in T2I generation. However, we observe that many $\langle$prompt, image$\rangle$ pairs still exhibit reasonably good prompt-image alignment, making them less informative for training a robust evaluator which requires exposure to diverse and fine-grained failure modes that capture subtle semantic, compositional, and visual inconsistencies. To address this limitation, we introduce a second-stage curation framework that selectively identifies informative prompt-image pairs that exhibit diverse failure modes. A large teacher VLM (\texttt{Qwen3-VL-235B}~\cite{Bai2025qwen}) is employed to perform structured evaluation of each $\langle$prompt, image$\rangle$ pair along two complementary dimensions: Text-to-Image Alignment (T2IA) and Image Quality Assessment (IQA). These evaluations are subsequently used for two purposes: (a) dataset curation, which when applied to GenDB yields \textbf{DynEvalInstruct}, a large-scale instruction dataset, and (b) providing fine-grained supervision for training a lightweight text-to-image evaluator. Now, we describe the structured evaluation framework used to obtain these annotations, followed by the curation strategy used to construct DynEvalInstruct.

\myparagraph{\textit{Structured Evaluation of a $\langle$prompt, image$\rangle$ pair.}} Given a $\langle$prompt, image$\rangle$ pair, we obtain T2IA and IQA scores through a structured intermediate process of QA generation followed by VQA-based scoring, as illustrated in \cref{fig:method_instruct}. This pipeline also serves as a knowledge distillation framework, transferring the evaluation capabilities of a large-scale teacher VLM to our lightweight evaluator. By decomposing evaluation into T2IA and IQA responses, we distill fine-grained supervision from the teacher model rather than relying on a single overall score. We next describe the QA generation and VQA-based scoring process in detail.

\noindent {\textbf{(i) Text-to-Image Alignment (T2IA).}}
In this stage, we evaluate how well a generated image aligns with its corresponding text prompt. 
\begin{itemize}[topsep=0pt, itemsep=0pt, partopsep=0pt, parsep=0pt, leftmargin=*]
\item[$\bullet$] \textbf{QA Generation:} 
Given a text prompt, the teacher VLM generates atomic binary `yes/no' questions to verify prompt alignment across objects, attributes, actions, spatial relations, scene consistency, \emph{etc}. It also generates complementary \textit{distortion-focused} questions targeting likely visual failures associated with the prompt (\emph{e.g.}, object fusion, floating objects, broken textures, and implausible structures). These are combined into a unified T2IA question set with corresponding target answers (\textbf{Steps 1a-1b} in \cref{fig:method_instruct}).

\item[$\bullet$] \textbf{VQA-based Scoring:} The combined question set and the image generated from the same prompt are then passed to the teacher VLM which assigns a score in $[1,5]$ for each question based on the given input rubric. The question-wise scores are averaged to obtain overall alignment score (\textbf{Step-2} in Fig.~\ref{fig:method_instruct}).

\end{itemize}

\noindent {\textbf{(ii) Image Quality Assessment (IQA).}}
While T2IA measures prompt faithfulness, the IQA branch evaluates the intrinsic visual quality of the generated image. Its goal is to identify whether the objects and relations present in the image are visually plausible, structurally coherent, and free from local artifacts or distortions. The text prompt is used only as a contextual reference during scene-graph construction, while the questions themselves are grounded in the visual content of the generated image.
\begin{itemize}[topsep=0pt, itemsep=0pt, partopsep=0pt, parsep=0pt, leftmargin=*]
\item[$\bullet$] \textbf{Scene Graph Generation:} 
Given an image, the VLM generates a scene graph with nodes representing objects and edges denoting relationships between them. To constrain generation, we provide the text prompt as a reference input, thereby implicitly reducing the likelihood of hallucinating objects and relationships not present in the image, compared to image-only VLM input counterparts (\textbf{Step-3a} in Fig.~\ref{fig:method_instruct}). 

\item[$\bullet$] \textbf{Image Quality QA Generation:}
For each node in the scene graph, the VLM decomposes the corresponding object into its constituent attributes and generates a set of binary `yes/no' questions targeting quality dimensions such as shape consistency, texture fidelity, completeness, structural realism, and 3D spatial cues (\emph{e.g.}, depth perception and relative distance). These questions, along with each of their corresponding `yes' or `no' answers, form the IQA question-answer set, enabling a detailed evaluation of object-wise and attribute-wise quality for each generated image (\textbf{Step-3b} in Fig.~\ref{fig:method_instruct}).

\item[$\bullet$] \textbf{VQA-based Scoring:}
We use the same scoring methodology as the VQA-based scoring in the T2IA pipeline, extending it to IQA questions for IQA scoring (\textbf{Step-4} in Fig.~\ref{fig:method_instruct}).

\end{itemize}

\noindent {\textbf{(iii) Combining T2IA and IQA Scores.}} For a given $\langle$prompt, image$\rangle$ pair, we first compute the T2IA and IQA scores and combine them via a weighted sum: $\alpha \times \text{(T2IA score)} + \beta \times \text{(IQA score)}$, where both $\alpha$ and $\beta$ are set to 0.5.

\myparagraph{\textit{Curation of DynEvalInstruct.}} Building upon the structured evaluation framework described above, we run the teacher VLM on each prompt-image pair in GenDB, yielding 500K $\langle \text{prompt}, \text{image}, \text{response}\rangle$ triplets, where each response contains structured T2IA and IQA outputs produced by the teacher model (for details, refer to Fig.~\ref{fig:method_instruct}). The resulting T2IA and IQA scores are then combined to obtain a single scalar score $S$. Since GenDB images originate from three model tiers with inherently different performance ranges, we use \textbf{tier-specific thresholds} $\delta_i$ (for $i \in \{1,2,3\}$). For a sample generated by a Tier-$i$ model, we mark it as a \emph{failure case} if $S < \delta_i$, and add it to a selected set $\mathcal{D}$. This yields a filtered subset of 250K triplets ($|\mathcal{D}| = 250\text{K}$), referred to as the \textbf{DynEvalInstruct} dataset. Notably, empirical analysis reveals that performance saturates at around 250K training samples, with diminishing returns beyond this point; accordingly, we construct DynEvalInstruct using 250K selected triplets (refer to Supp. \cref{tab:data_scaling}). The resulting dataset maintains balanced representation across prompt tiers while ensuring coverage of challenging semantic, compositional, and visual failure cases across the full spectrum of model capabilities. Further details on the thresholds are provided in the Supp. \cref{supp:tst}.

\begin{table}[!t]
\centering
\setlength{\tabcolsep}{2pt}
\renewcommand{\arraystretch}{1.15}
\caption{\textbf{Zero-shot quantitative comparison between DynEval and existing methods for predicting overall alignment scores across multiple benchmarks.} Columns correspond to different datasets containing ⟨prompt, image⟩ pairs with associated human ratings, while rows represent evaluation methods used to estimate alignment scores. Notably, \textbf{DynEval} evaluators are trained entirely without human annotations yet achieve strong agreement with human judgments across diverse evaluation settings. GenEval~\cite{GenEval} scores are reported only for annotation-compatible datasets; unavailable EvalMuse~\cite{EvalMuse-40k} entries are left blank. Best: \hcolor{\large$\bullet$}, Second-best: \blue{\large$\bullet$}.}
\label{tab:main_comp}
\vspace{-3mm}
\resizebox{\textwidth}{!}{
\begin{tabular}{
>{\raggedleft\arraybackslash}p{3.5cm}
|c c
|c c
|c c
|c c
|c c
|c c
|c c
}
\toprule

\multirow{2}{*}{\textbf{Evaluation Method}}
& \multicolumn{2}{c|}{\textbf{EvalMuse}~\cite{EvalMuse-40k}}
& \multicolumn{2}{c|}{\textbf{GenAI-B}~\cite{lin2024vqascore}}
& \multicolumn{2}{c|}{\textbf{TIFA}~\cite{tifa}}
& \multicolumn{2}{c|}{\textbf{RichHF}~\cite{liang2024rich}}
& \multicolumn{2}{c|}{\textbf{GenEval}~\cite{GenEval}}
& \multicolumn{2}{c|}{\textbf{EvalMi}~\cite{lmm4lmm_evalmi-50k}}
& \multicolumn{2}{c}{\textbf{T2I-Eval-B}~\cite{t2i_eval_bench}}
\\

\cline{2-15}
& SRCC & PLCC
& SRCC & PLCC
& SRCC & PLCC
& SRCC & PLCC
& SRCC & PLCC
& SRCC & PLCC
& SRCC & PLCC
\\

\midrule
    CLIPScore~\cite{hessel2021clipscore} & 0.2993 & 0.2933 & 0.1676 & 0.2030 & 0.3003 & 0.3086 & 0.0570 & 0.3024 & 0.0654 & 0.0985 & 0.2607 &	0.3072 & 0.0678	& 0.1260\\
    BLIPv2Score~\cite{li2023blip} & 0.3583 & 0.3348 & 0.2734 & 0.2979 & 0.4287 & 0.4543 & 0.1425 & 0.3105 & 0.2669 & 0.3167 & 0.2900 &	0.3468 & 0.1223 & 0.1321\\
    
    ImageReward~\cite{imagereward} & 0.4655 & 0.4585 & 0.3400  & 0.3786 & 0.6211 & 0.6336 & 0.2747 & 0.3291  & 0.4149 & 0.4603 & 0.4991 & 0.5523 & 0.1349 & 0.1144\\
    
    PickScore~\cite{kirstain2023pick} & 0.4399 & 0.4328 & 0.3541 & 0.3631 & 0.4279 & 0.4342 & 0.3916 & 0.4133 & 0.2787 & 0.3015 & 0.4611 & 0.4692 & 0.2375 & 0.2440\\
    
    HPSv2~\cite{wu2023human} & 0.3745 & 0.3657 & 0.1371 & 0.1693 & 0.3647 & 0.3804 & 0.1871 & 0.2577 & 0.3223	& 0.3580 & 0.5336 & 0.5525 & 0.2916 & 0.2990\\
    
    VQAScore~\cite{lin2024vqascore} & 0.4877 & 0.4841 & 0.5534 & 0.5175 & 0.6951 & 0.6585 & 0.4826 & 0.4094 & 0.5057 & 0.5040 & 0.6062 &	0.6118 & 0.3348 & 0.3570\\
    
    GenEval~\cite{GenEval} & - & - & - & -&-&-& - & - & 0.4753 & 0.4509 & - & - & - & -\\
    
    FGA-BLIP2-OS~\cite{EvalMuse-40k} & \blue{0.7742} & \blue{0.7722} & \blue{0.5637} & \blue{0.5673} & \blue{0.7604} & \blue{0.7442} & 0.5123 & 0.5455 & 0.5170 & 0.5278 & 0.6724 & 0.6945 & 0.3582 & 0.3471\\

    TIIF-Bench~\cite{tiff_bench} & - & - & 0.4387 & 0.4020 & 0.6007 & 0.5776 & 0.4342 & 0.4292 & 0.5600 & 0.5447  & 0.4474 & 0.4256 & 0.3297 & 0.3016\\
    T2I-Eval-Bench~\cite{t2i_eval_bench} & - & - & 0.4996 & 0.4773 & 0.6011 & 0.6246 & 0.5033 & 0.4990 & 0.5847 & 0.5610 & 0.4258 & 0.4075 & 0.3744 & 0.3629\\
    UniGenBench++~\cite{unigenbench++} & - & - & 0.4605	& 0.4441 & 0.6078 & 0.5892 & 0.5333	& 0.5172 & 0.5661 & 0.5494 & 0.4611 & 0.4554 & 0.3971 & 0.3818\\
    T2I-CoReB (8B)~\cite{T2I-CoReBench} &  - & - & 0.4614 & 0.4414 & 0.6126	& 0.6112 & \blue{0.6024} & \blue{0.5809} & 0.5626 & 0.5810 & 0.5583 & 0.5414 & 0.4035 & 0.3879\\
    GenEval 2 (8B)~\cite{geneval2} & - & - &  0.4817 & 0.4571 & 0.6824 & 0.6734 & \hcolor{0.6311} & \hcolor{0.6021} & \hcolor{0.6044} & \hcolor{0.6175} & \blue{0.7245} & \blue{0.7098} & 0.4222 & 0.4172\\
    LongT2IBench~\cite{yang2026longt2ibench} & - & - & 0.3747 & 0.3763 & 0.6775 & 0.6922 & 0.5174 & 0.5161 & 0.4544 & 0.5999 & 0.5056 & 0.5485 & \textbf{\blue{0.4781}} & \textbf{\blue{0.4563}}\\
    
    \midrule
    
    \rowcolor{lightblue} \textbf{DynEval-2B (Ours)} & 0.7765 & 0.7798 & 0.5715 & 0.5695 & 0.7827 & 0.7715 & 0.5226 & 0.5545 & 0.5668 & 0.5820 & 0.7346 & 0.7156 & 0.4535 & 0.4410\\

    \rowcolor{lightblue} \textbf{DynEval-4B (Ours)} & \hcolor{0.7932} & \hcolor{0.7945} & \hcolor{0.5945} & \hcolor{0.5817} & \hcolor{0.8022} & \hcolor{0.8034} & 0.5457 & 0.5691 & \blue{0.5894} & \blue{0.6073} & \hcolor{0.7944} & \hcolor{0.7889} & \hcolor{0.4857} & \hcolor{0.4636}\\
    \bottomrule
    \end{tabular}
  }
\end{table}

\section{DynEVAL}
\label{sec:dyneval}

Training an evaluator for T2I models at scale requires reliable data which is infeasible to obtain through human annotation for hundreds of thousands of $\langle$prompt, image$\rangle$ pairs. We therefore adopt a distillation-based approach, where the capabilities of a strong teacher VLM are transferred to a smaller evaluator model. We propose \textbf{DynEval}, with two variants: DynEval-2B and DynEval-4B, trained by distilling responses from \texttt{Qwen3-VL-235B}~\cite{Bai2025qwen}. A key design choice is that we do not distill DynEval as a black-box regressor that directly maps a $\langle$prompt, image$\rangle$ pair to a single score. Importantly, VLMs tend to produce redundant questions when directly asked to generate questions conditioned on either text prompt (\textbf{T2IA} process) or the image (\textbf{IQA} process), which violates atomicity by repeatedly verifying overlapping attributes and relations, thereby artificially inflating the final evaluation score. Motivated by~\cite{cho2023davidsonian, EvalMuse-40k}, we also emphasize \textit{atomicity} and \textit{coverage}: questions should be atomic, avoid containing hallucinations, and maintain valid dependencies. To implement this structured distillation, we introduce three \textbf{task-specific tokens} that explicitly trigger different evaluation procedures during training and inference (refer to~\cref{fig:method_instruct}).

\begin{itemize}[topsep=0pt, itemsep=0pt, partopsep=0pt, parsep=0pt, leftmargin=*]
\item[$\bullet$] \textbf{\texttt{<T2IA>}} triggers generation of diverse binary verification questions using only the text prompt. With this token, the model also generates distortion-checking questions with a separate instruction-tuning prompt.

\item[$\bullet$] \textbf{\texttt{<IQA>}} activates scene-graph generation and fine-grained image-quality assessing question generation with respective instruction-tuning prompt. 

\item[$\bullet$]  \textbf{\texttt{<EVALUATION>}} triggers VQA-style answering and scoring. Given a generated image and a set of questions produced by the \texttt{<T2IA>} or \texttt{<IQA>} token, the model answers each question and outputs a score in the range $[1,5]$. 
\end{itemize}

\noindent These objectives are trained in a curriculum: the evaluator first learns to generate structured questions and then learns to answer them. This reduces model collapse during fine-tuning and stabilizes the evaluator’s reasoning behavior. In practice, we find that the tokenized curriculum helps the model disentangle \emph{semantic alignment}, \emph{visual integrity}, and \emph{relative preference}--three crucial dimensions that previous evaluators either conflated or ignored. During inference, DynEval first triggers \texttt{<T2IA>} to generate prompt-grounded verification questions and distortion checks, followed by \texttt{<EVALUATION>} to obtain the T2IA score. Next, DynEval triggers \texttt{<IQA>} to generate a scene graph and image-quality assessing questions sequentially, followed by \texttt{<EVALUATION>} to obtain the IQA score. Together, these elements allow \textbf{DynEval} to dynamically adapt to new prompts without requiring pre-defined questions~\cite{EvalMuse-40k}, attributes~\cite{tifa, geneval2}, or ground-truth labels~\cite{GenEval}, while producing judgments closely aligned with human evaluators.

\section{Results and Discussions}

\subsection{Experimental Setup}

\myparagraph{\textit{Training Settings.}} We do full-scale fine-tuning of \texttt{Qwen3-VL-4B} to obtain \textbf{DynEval-4B} and \texttt{Qwen3-VL-2B} to obtain \textbf{DynEval-2B}. The models are trained using the \textbf{DynEvalInstruct} dataset consisting of 250K samples, where each sample contains an image, a text prompt, and a detailed response generated through the T2IA and IQA pipelines (refer to \cref{fig:method_instruct}). For training, we use a linear warmup followed by cosine decay, with a peak learning rate of $2\times10^{-5}$. The models are trained for one epoch on 4x NVIDIA H200 GPUs.

\myparagraph{\textit{Evaluation Settings.}} We compare DynEval with \textbf{14} existing evaluation methods on \textbf{11} benchmarks by reporting the Spearman Rank Correlation Coefficient (\textbf{SRCC}) and Pearson Linear Correlation Coefficient (\textbf{PLCC}) to measure the correlation between the evaluator predicted score and human-annotated score. Higher correlation values indicate stronger agreement with human judgment. We use results reported in the original evaluator papers whenever available; otherwise, we evaluate official checkpoints on newer challenging benchmarks. \cref{tab:main_comp} reports comparative results on the \textbf{7} benchmarks that offer publicly available prompts, generated images, and human annotations, while the additional benchmarks in \cref{tab:added_cross_benchmark_results} lack human annotations. Detailed statistics of the considered evaluation benchmarks are provided in Supp. \cref{supp:dataset_Details}.

\subsection{Experimental Results.}

\myparagraph{\textit{Qualitative Results.}} In Fig.~\ref{fig:qualitative_comp}, we showcase a comparison between prediction scores of \textbf{DynEval-4B} with existing evaluation methods~\cite{GenEval,DPGBench,EvalMuse-40k,tifa} on representative image-text pairs. As shown, popular T2I evaluations, \emph{e.g.}, GenEval~\cite{GenEval} and DPG-Bench~\cite{DPGBench}, often assign high scores to distorted images despite their low human ratings. Similarly, methods like EvalMuse~\cite{EvalMuse-40k} and TIFA~\cite{tifa} also exhibit substantial deviations from human judgments. In contrast, our proposed DynEval produces scores that are consistently more aligned with human evaluation, demonstrating its effectiveness in jointly assessing image-text alignment and image quality. Further qualitative results are shown in Supp. \cref{fig:supp_teaser_fig}.

\begin{wraptable}[13]{r}{0.65\columnwidth}
\centering
\vspace{-8mm}
\setlength{\tabcolsep}{2pt}
\renewcommand{\arraystretch}{1.2}
\caption{\textbf{Additional zero-shot comparison between DynEval and recent evaluation methods.} Columns denote benchmark datasets and rows denote evaluators. Best: \hcolor{\scalebox{1.1}{$\bullet$}}; second-best: \blue{\scalebox{1.1}{$\bullet$}}.}
\label{tab:added_cross_benchmark_results}

\resizebox{0.65\textwidth}{!}{
\begin{tabular}{
>{\raggedleft\arraybackslash}p{3.5cm}
|c c
|c c
|c c
|c c
}
\toprule

\multirow{2}{*}{\textbf{Evaluation Method}}
& \multicolumn{2}{c|}{\textbf{GenEval 2}~\cite{geneval2}}
& \multicolumn{2}{c|}{\textbf{TIIF-B}~\cite{tiff_bench}}
& \multicolumn{2}{c|}{\textbf{UniGenB}++~\cite{unigenbench++}}
& \multicolumn{2}{c}{\textbf{T2I-CoReB}~\cite{T2I-CoReBench}}
\\

\cmidrule(lr){2-3}
\cmidrule(lr){4-5}
\cmidrule(lr){6-7}
\cmidrule(lr){8-9}

& SRCC & PLCC
& SRCC & PLCC
& SRCC & PLCC
& SRCC & PLCC
\\

\midrule

VQAScore~\cite{lin2024vqascore}
& 0.6223 & 0.5208
& 0.3728 & 0.3353
& 0.3109 & 0.2991
& 0.2469 & 0.2668
\\

FGA-BLIP2-OS~\cite{EvalMuse-40k}
& 0.3868 & 0.3750
& 0.3581 & 0.4437
& 0.3521 & 0.3457
& 0.3154 & 0.3075
\\

TIIF-Bench~\cite{tiff_bench}
& 0.3432 & 0.3322
& 0.3917 & 0.3690
& 0.3254 & 0.3111
& 0.3602 & 0.3562
\\

T2I-Eval-Bench~\cite{t2i_eval_bench}
& 0.3904 & 0.3862
& 0.3392 & 0.3190
& 0.3473 & 0.3356
& 0.3811 & 0.3767
\\

UniGenBench++~\cite{unigenbench++}
& 0.4602 & 0.4072
& 0.4788 & 0.4553
& 0.3888 & 0.3752
& 0.3671 & 0.3454
\\

T2I-CoReB (8B)~\cite{T2I-CoReBench}
& 0.4813 & 0.4955
& 0.5196 & 0.5671
& 0.4004 & 0.3849
& \textbf{\blue{0.4469}} & \textbf{\blue{0.4352}}
\\

GenEval 2 (8B)~\cite{geneval2}
& \textbf{\blue{0.7120}} & \textbf{\blue{0.7017}}
& \textbf{\blue{0.5495}} & \textbf{\blue{0.6028}}
& \textbf{\blue{0.4102}} & \textbf{\blue{0.3858}}
& 0.4435 & 0.4315
\\

\midrule

\rowcolor{lightblue}
\textbf{DynEval-4B} \textbf{(Ours)}
& \textbf{\hcolor{0.7508}} & \textbf{\hcolor{0.7132}}
& \textbf{\hcolor{0.5980}} & \textbf{\hcolor{0.6298}}
& \textbf{\hcolor{0.4958}} & \textbf{\hcolor{0.5042}}
& \textbf{\hcolor{0.4856}} & \textbf{\hcolor{0.4716}}
\\

\bottomrule
\end{tabular}
}

\end{wraptable}
\myparagraph{\textit{Quantitative Results.}} As shown in Tabs.~\ref{tab:main_comp} and \ref{tab:added_cross_benchmark_results}, \textbf{DynEval-4B} achieves SOTA performance on 9 out of 11 benchmarks and ranks second best on one additional benchmark. This demonstrates strong generalization across diverse text-to-image evaluation settings. Averaged over all benchmarks in Tabs.~\ref{tab:main_comp}-\ref{tab:added_cross_benchmark_results}, DynEval-4B surpasses the previous best evaluators by a relative margin of \textbf{+4.77\%}, with the relative improvement increasing to \textbf{+7.61\%} on the benchmarks where it establishes a new state of the art in terms of SRCC. Notably, DynEval-4B achieves the largest relative gains on EvalMi~\cite{lmm4lmm_evalmi-50k} (+9.65\%) and GenAI-Bench~\cite{li2024genai} (+5.46\%) in \cref{tab:main_comp}, while attaining substantial relative improvements on UniGenBench++~\cite{unigenbench++} (+20.87\%) and TIIF-Bench~\cite{tiff_bench} (+8.83\%) in \cref{tab:added_cross_benchmark_results}. Furthermore, on the RichHF and GenEval benchmarks, DynEval-4B trails the larger 8B evaluators T2I-CoReBench~\cite{T2I-CoReBench} and GenEval2~\cite{geneval2} by relative margins of 9.41\% and 2.48\%, respectively, while only using half the parameters of these methods. Nonetheless, DynEval-4B achieves the strongest overall performance across the complete suite of evaluation benchmarks. These results highlight the effectiveness and robustness of DynEval as a unified and highly generalizable T2I model evaluation framework.

\myparagraph{\textit{Model and Data Scaling.}} Furthermore, scaling our evaluator from 2B to 4B parameters consistently improves performance across all benchmarks in \cref{tab:main_comp}, yielding an average relative SRCC gain of \textbf{4.62\%}. The largest improvements occur on EvalMi (\textbf{+8.14\%}) and T2I-Eval-Bench (\textbf{+7.10\%}), while even benchmarks with smaller gains such as EvalMuse (\textbf{+2.15\%}) show consistent improvement. This demonstrates that DynEval benefits from model scaling. We further provide ablations on teacher model selection (\cref{tab:teacher_ablation}) and fine-tuning dataset scaling (\cref{tab:data_scaling}), analogous to model scaling, in the Supp.

\begin{wraptable}[8]{r}{0.62\columnwidth}
\small
\centering
\vspace{-10mm}
\caption{Statistics and quality analysis of IQA questions generated by \textbf{DynEval-4B} across 11 evaluation datasets, using \texttt{Qwen3-VL-235B} as the teacher.}

\resizebox{0.6\columnwidth}{!}{%
\begin{tabular}{l||c|c|c|c||c|c}
\hline
\# Objects & \multicolumn{4}{c||}{Question Count Statistics }& \multicolumn{2}{c}{Quality of IQA Generated Questions}\\
\cline{2-7}
in Image & Mean & Median & Min & Max & Coverage Ratio $\uparrow$ & BERTScore $\uparrow$ \\
\hline
 
1--5   & 14.2  & 14 & 5 & 25 & \cellcolor{lightblue}0.885 & \cellcolor{lightblue}0.899  \\

\cellcolor{lightblue!27}6--10 & \cellcolor{lightblue!27}32.2 & \cellcolor{lightblue!27}33 & \cellcolor{lightblue!27}24 & \cellcolor{lightblue!27}40 & \cellcolor{lightblue!51}0.824 & \cellcolor{lightblue!51}0.871 \\

\cellcolor{lightblue!51}11--15 & \cellcolor{lightblue!51}47.6 & \cellcolor{lightblue!51}45 & \cellcolor{lightblue!51}36 & \cellcolor{lightblue!51}61 & \cellcolor{lightblue!27}0.752 & \cellcolor{lightblue!27}0.817 \\

\cellcolor{lightblue}16--20 & \cellcolor{lightblue}66.6 & \cellcolor{lightblue}63 & \cellcolor{lightblue}47 & \cellcolor{lightblue}100 & 0.696 & 0.772\\

\hline
\end{tabular}%
}
\label{tab:question_stats}
\end{wraptable}
\myparagraph{\textit{Statistics on Dynamic Question Generation.}}
In Tab.~\ref{tab:question_stats}, we validate the dynamic IQA question generation process by analyzing the number and quality of questions generated by \textbf{DynEval-4B} across 11 evaluation datasets. It is observed that the number of questions scales almost linearly with the number of objects in the image. In addition to reporting question-count statistics, we also evaluate the quality of the generated IQA questions independently of the final DynEval score. Specifically, using \texttt{Qwen3-VL-235B}~\cite{Bai2025qwen} as the teacher VLM, we compute: \textbf{(i) Coverage Ratio}, defined as the IoU between teacher and student response sets; and \textbf{(ii) BERTScore similarity}~\cite{zhang2019bertscore}, to measure semantic alignment between teacher-generated and DynEval-generated questions, both reported in the range $[0,1]$.

\begin{figure}[!t]
    \centering
    \includegraphics[width=\textwidth]{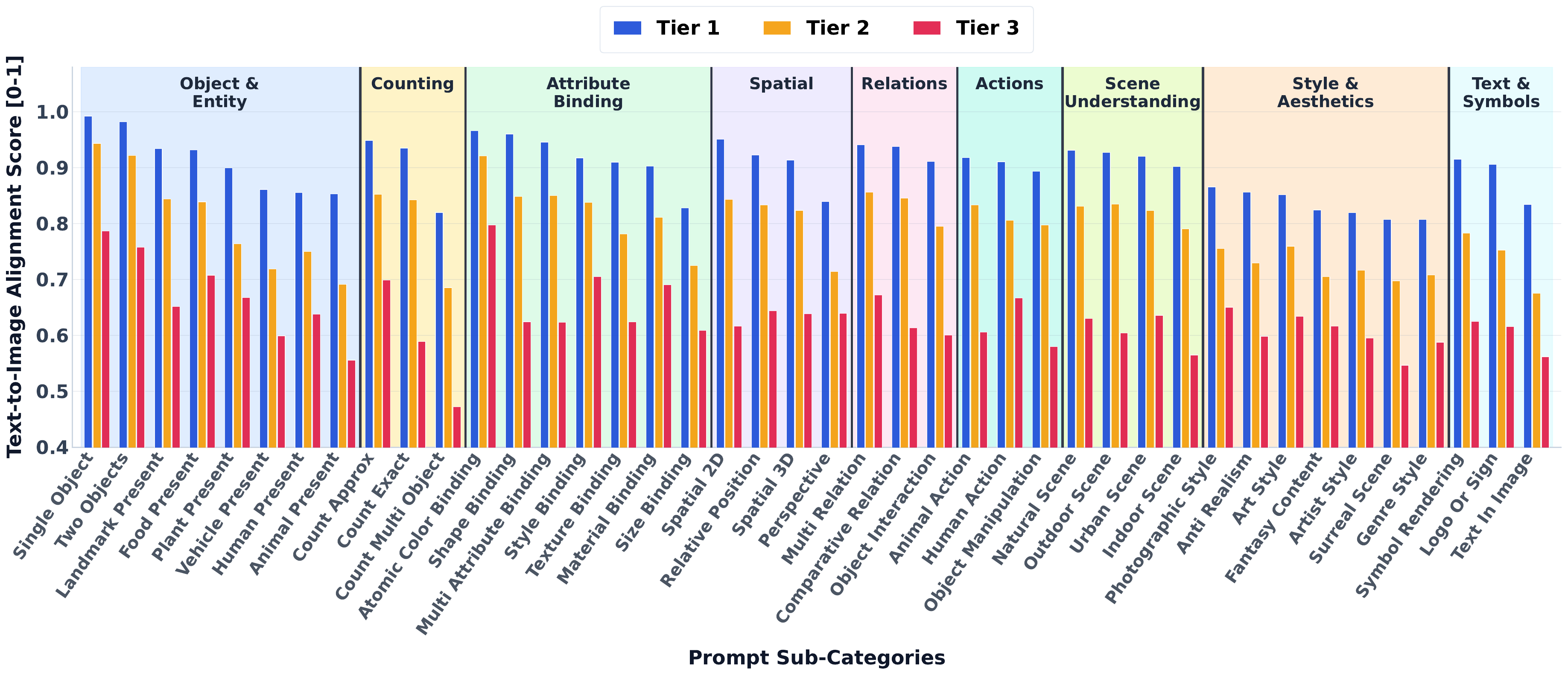}
    \vspace{-7mm}
    \caption{\textbf{Fine-grained capability analysis of 36 T2I models on the DynEval-1K evaluation set using DynEval-4B across 42 prompt subcategories and 9 semantic dimensions.} While Tier-1 models exhibit stronger alignment across most categories, all tiers show substantial performance degradation on challenging aspects such as \textit{count multi objects}, \textit{style and aesthetics}, \textit{human present}, and \textit{size binding}. These findings indicate that, despite significant progress in object grounding and attribute binding, these capabilities remain persistent challenges for current T2I models.}
    \label{fig:category_score}
    \vspace{-5mm}
    
\end{figure}

\subsection{Understanding T2I Models' Failure Attributes}
\label{sec:failure}

As shown in \cref{fig:category_score}, while modern T2I models demonstrate strong capabilities in basic object grounding, scene understanding, and standard spatial relations and reasoning, several challenges persist across all capability tiers, which can be summarized as follows: 
\textbf{(i)} object counting emerges as one of the most challenging capabilities, with performance degrading rapidly as compositional complexity increases with the number of objects and their attributes.
\textbf{(ii)} Prompts involving humans and animals continue to be significantly more difficult than those involving inanimate entities (\emph{e.g.}, landmarks, food, and vehicles), highlighting challenges in modeling fine-grained structures and appearances. 
\textbf{(iii)} Similarly, models struggle with complex attribute binding, particularly under counterfactual or uncommon size relationships that contradict real-world priors, suggesting a strong reliance on learned visual priors rather than robust relational reasoning. \textbf{(iv)} Perspective-dependent prompts involving diverse camera viewpoints remain challenging even for the most capable models, highlighting limitations in geometric understanding and viewpoint control.
\textbf{(v)} Accurate text and symbol generation also remains an unresolved challenge, with models frequently failing to faithfully preserve textual content within generated images. Although higher-capability models achieve better overall alignment, the relative difficulty ordering of these challenging dimensions remains largely consistent across model tiers. To further analyze this behavior, we provide comparisons between the best and worst models within each tier across all 42 evaluation sub-categories in the Supp. \cref{supp:failure}.

\begin{wrapfigure}[18]{r}{0.36\linewidth}    
    \centering
    \vspace{-8mm}
    \includegraphics[width=0.98\linewidth]{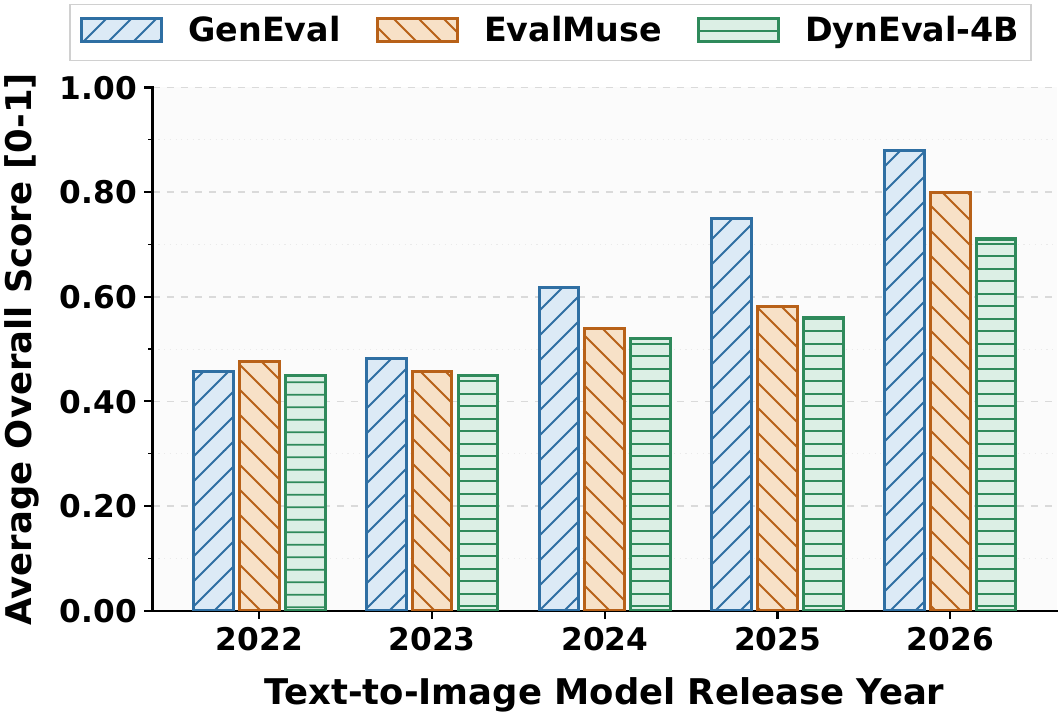}
    \vspace{-6mm}
    \caption{
    Average scores assigned by existing methods~\cite{EvalMuse-40k, GenEval} and DynEval-4B to T2I models grouped by publication year. While existing evaluators show steadily increasing scores for newer models, DynEval-4B produces more calibrated assessments by jointly evaluating text-image alignment and image quality (visual realism).
    }
    \label{fig:TREND}
    
\end{wrapfigure}

Despite rapid advances in text-to-image (T2I) generation, reliable evaluation remains an open challenge. As illustrated in Fig.~\ref{fig:TREND}, recent T2I models achieve steadily improving scores on benchmarks such as GenEval~\cite{GenEval}. However, as a detector-based framework, GenEval mainly rewards object and attribute presence, such that performance gains may reflect improved attribute satisfaction rather than genuine advances in visual fidelity. Although recent VQA-based evaluators improve semantic assessment, they remain largely alignment-centric and overlook fine-grained perceptual quality~\cite{EvalMuse-40k}. In contrast, our approach jointly evaluates text-image alignment and image quality, yielding a more holistic assessment and substantially stronger agreement with human judgments.
\section{Conclusion}

In this work, we presented \textbf{DynEval}, a holistic and truly dynamic evaluation framework for text-to-image models that jointly assesses \textit{text-to-image alignment} (T2IA) and \textit{image quality} (IQA), targeting failure modes that are often missed or inconsistently scored by existing automated evaluators. To enable scalable training without manual annotations, we constructed \textbf{GenDB}, a 500K prompt-image dataset, and subsequently derived from it \textbf{DynEvalInstruct}, a 250K instruction dataset distilled from a teacher VLM. Leveraging this data, we trained lightweight evaluators (DynEval-2B/4B) that achieve overall state-of-the-art performance across \textbf{11} benchmarks while also providing fine-grained diagnostic insights into persistent weaknesses of modern T2I models.

\vspace{-2mm}
\section*{Acknowledgements}
\vspace{-2mm}

We gratefully acknowledge Kotak-IISc AI/ML Centre (KIAC) for the generous conference travel grant and the GPU resources that supported this research.

\bibliographystyle{splncs04}
\bibliography{main}

\input{supp}
\end{document}